\documentclass[11pt,journal,compsoc]{IEEEtran}
\usepackage{blindtext, graphicx}
\usepackage{subcaption}
\usepackage{booktabs}
\usepackage[table,xcdraw]{xcolor}
\usepackage[justification=centering]{caption}
\usepackage{flushend}
\usepackage{tikz}
\def\checkmark{\tikz\fill[scale=0.4](0,.35) -- (.25,0) -- (1,.7) -- (.25,.15) -- cycle;}

\begin{document}

\title{Using Data Analytics to Detect Anomalous States in Vehicles}

\author{Sandeep Nair Narayanan, Sudip Mittal \& Anupam Joshi\\ \{sand7, smittal1, joshi\}@umbc.edu \\ Department of Computer Science \& Electrical Engineering\\University of Maryland Baltimore County\\Baltimore, 21227 \\ Maryland, U.S.A.}

%\institute{Department of Computer Science \& Electrical Engineering\\University of Maryland Baltimore County\\Baltimore, 21227 \\ Maryland, U.S.A.}

\maketitle

\begin{abstract}
Vehicles are becoming more and more connected, this opens up a larger attack surface which not only affects the passengers inside vehicles, but also people around them. These vulnerabilities exist because modern systems are built on the comparatively less secure and old CAN bus framework which lacks even basic authentication. Since a new protocol can only help future vehicles and not older vehicles, our approach tries to solve the issue as a data analytics problem and use machine learning techniques to secure cars. We develop a Hidden Markov Model to detect anomalous states from real data collected from vehicles. Using this model, while a vehicle is in operation, we are able to detect and issue alerts. Our model could be integrated as a plug-n-play device in all new and old cars. 
\end{abstract}

\section{Introduction}
According to US department of transportation~\cite{unitedstatesdepartmentoftransportation}, 88\% of all people are drivers with 1.9 mean number of vehicles per household in the US~\cite{adamverbach}. Vehicles have become an integral part of our life and automobile technology has progressed to address various needs. Earlier, it was the sole responsibility of the driver to control various activities in a vehicle, but with the proliferation of micro-controllers and security features many such tasks have been delegated to electronic chips. The controllers have started taking precursory actions. To ease inter-controller communication, a common internal communication bus was introduced instead of interconnecting them separately. The design of the bus gave more importance to performance than security. Hence the protocol lacked any kind of authentication schemes or non-repudiation mechanics.  

Different research enlists multiple attacks against a vehicle. Hoppe et al.~\cite{hoppe2008security} describe different attacks which are possible by malicious modifications of ECU code. Using one such modification they were able to open the car windows~\cite{hoppe2007sniffing} automatically when it reaches the speed of 200 kilometer per hour. These vulnerabilities are severe since they directly affect passenger safety. In another instance, they hacked the comfort control unit so as to control the warning lights. Since the comfort control ECU is responsible anti-theft functionality, they could have easily disabled the alarm system to ease unauthorized entry to a vehicle. Researchers were also able to manipulate the ‘Airbag Control Systems’~\cite{hoppe2008security} in the vehicle using a similar technique. Koscher et al.~\cite{koscher2010experimental} experimentally evaluated and demonstrated various problems with the underlying system structure. They illustrated how easy it was for a determined attacker to infiltrate various ECU’s and circumvent different security systems. They were able to attack various things like the speedometer, lights, brakes, doors locks etc.

Recently researchers were able to exploit the weakness in a car system build on top of a basic CAN network by injecting malicious data into the internal bus. The effects of their hack ranged from acts like switching the lights on, to potentially fatal acts like applying brakes. Charlie Miller and Chris Valasek demonstrated a hack by exploiting the vulnerability by attaching an external device~\cite{miller2013adventures}. They demonstrated their recent work at Blackhat 2015, in which they hacked a Jeep Cherokee~\cite{miller2015blackhat} remotely without even attaching any external device. More reports~\cite{vwhack} have come out listing various vehicles from popular car makers like Volkswagen, Skoda, Volvo, etc. that are vulnerable to another kind of crypto attack on key less entry.

We first collected data from different vehicles and formulated the problem into a data analytic problem. Then we used HMM to create a model. Once the model is generated, we use it to predict  any unsafe or anomalous states from the data flowing on the CAN bus. We evaluated our system by generating multiple anomalous scenarios by logically modifying the collected real data.

\subsection{Background}
Older cars were just mechanical devices. But modern vehicles can be considered as a collection of Electronic Control Units (ECU), Sensor and actuators. The different control units include Anti-lock Brake System (ABS), Adaptive Cruise control, Active Suspension, Active Vibration Control, Entertainment System, Lane Keeping Assist, Electronic Power Steering, Adaptive Front lighting etc. Many of these systems depend on each other and there is a requirement to interconnect them. Since interconnecting each of them separately is not efficient, Bosch proposed CAN bus, with the most recent CAN specification ‘CAN 2.0’ published in the year 1991~\cite{boschsemiconductors}. In the year 1993 the International Organization for Standardization released the CAN standard ISO 11898~\cite{iso118981:2003}. It is designed to be fast and simple in architecture.

\section{Related Work}

There are two ways to address the current security problem. One way is to prevent the attack from happening and the second one is to detect and mitigate the potential risk. One of the main reasons which enable attackers to inject potentially malicious messages on the CAN bus is that the protocol lacks authentication mechanisms~\cite{koscher2010experimental}. Researchers tried to address this problem by using cryptography. Wolf et al.~\cite{wolf2007state} looked at the requirements of cryptographic functions for car security. They proposed embedded solutions to add cryptographic functions to different ECU’s which can provide security against malicious manipulations.

\subsection{Attack Prevention}

Hazem and Fahmy~\cite{hazem2012lcap} proposed LCAP, a light weight CAN authentication protocol to secure the CAN bus in which they reduced communication overhead and computational complexity associated with cryptography. In this protocol, they used a ‘one-way hash function’ to generate a ‘magic number’ which is selected by the sender and can only be verified by the receiver. The magic number would be sent either using the ‘extended identifier field’ or as a payload over the CAN bus. LCAP requires a predefined key and various session keys to be exchanged between each sender and receiver, every time in the initial phase of the protocol. 

Another authentication protocol is CANAuth~\cite{van2011canauth} by Herrewege et al. This protocol is backward compatible with the currently used protocol. Apart from proposing a new protocol, they identified different restrictions on the CAN bus system like hard real-time constraints, message length restriction, lack of bi-directional communication, etc. In this protocol, authentication data is transmitted out of band which provides a maximum length of 15 bytes for the authentication message. After key exchange, the authentication process will have two parts, a counter value to prevent replay attacks and 80 LSB of HMAC of the counter value and CAN message data. This protocol also requires a pre-shared key.

LiBrA-CAN~\cite{grozalibra} by Groza et al. is yet another protocol to secure the CAN bus. Instead of providing independent authentication for each sensor or ECU, they assigned keys for a group of these devices. They employed ‘key splitting’ and ‘MAC mixing’ in this protocol to provide security. Apart from the basic authentication scheme, they also discussed several variations of their protocol which are broadly classified as ‘master orient authentication schemes’ and ‘distributed authentication schemes’. The master oriented authentication schemes include centralized, cumulative and load balanced authentication schemes, while the distributed authentication scheme includes alternatives like two stage authentication and multi master authentication schemes.

\subsection{Attack Detection and Mitigation}

Koscher et al.~\cite{checkoway2011comprehensive} discusses the importance of ‘detection mechanisms’ versus ‘prevention mechanisms’. They solidify our hypothesis that, operational and economic realities in the domain demands a detection strategy unlike the prevention strategies using cryptography mentioned above. Ruta et al.~\cite{ruta2010mobile} collected the OBD data from the CAN bus and analyzed it to infer potential risk factors and provide the user with warning. In their method they fused the OBD speed and RPM information with external data like weather information, location information, etc. using simple ‘data fusion’ algorithms to perform logic based matchmaking. They inferred road and traffic conditions, driving behavior, etc. and generated suggestions to minimize risk factors. For example, their system suggests the driver to use ABS and fog lamps along with slow driving if it detects a foggy weather at a particular location.

\subsection{Hidden Markov Models and their Applications}
A Markov process is a stochastic model that follows the Markov property, which states that any new state transition depends only on the current state. A Hidden Markov Model is similar to a Markov process but with hidden or unobserved states. Each of these hidden states will be associated with a set of observations. To create a HMM model, we generate two set of probabilities, Transition probabilities and Emission probabilities. Transition probability controls how a new state, lets say ``S(t)'', is chosen from a current state ``S(t-1)''. The emission probability is probability that a specific set of observations will be generated given current hidden state ``S(t)''. During model generation we try to estimate these probabilities using the given data set. The different problems solved using HMM models include the following. Given the model, what is the probability of an observation sequence, what is the most likely sequence of hidden states corresponding to an observation sequence, etc.

In our project we use this Hidden Markov Models which are quite popular for analyzing time series data.  et al.[19] discuss three time series clustering approaches: raw-data-based,  feature-based,  model-based. They also suggest various methods to formulate clusters given a time series dataset. They advise starting with feature extraction,  followed by clustering which will produce Clusters and maybe Cluster Centers. 

Hidden markov models have been applied to many problems in various fields like finance, bioinformatics, etc. Ziv et al.~\cite{bar2004analyzing} successfully apply it to analyze time series gene expression data so as to study a wide range of biological systems. They stress that hidden markov models help them to infer causality from the temporal response pattern and address the challenge of handling different non-uniform sampling rates. Zhang et al..~\cite{zhang2004prediction} use hidden markov models to analyze financial data. They take historical multidimensional and complex nonlinear data from various financial indexes and develop a hidden markov prediction system to find a possible future value of a stock price.  

An interesting paper by Guo et al.~\cite{guo2009using} that is quite relevant to our study uses accelerometer, GPS data to develop a movement and behavior model for cattle by using hidden markov models. The authors collect real data for individual cows in the herd and then try to predict their movements using machine learning models. The authors develop a 3 state model which was able to describe animal movement and state transition behavior accurately.  

As described above Koscher et al.\cite{checkoway2011comprehensive}  described the importance of a detection scheme due to practical issues. Moreover the data analytics on OBD data~\cite{ruta2010mobile} proves useful. We believe that a machine learning approach can provide us a method to detect abnormal car behaviour using data from the OBD port. Guo et al.~\cite{guo2009using} have used machine learning algorithms like hidden markov models in a similar manner. This leads us to believe that a similar model can detect abnormal states from OBD data. 

In our project we encounter a regular stream of OBD messages which are transmitted on a vehicle’s CAN bus. In order to determine if the car is in normal or abnormal mode of operation we analyzed this regular stream of OBD messages. We then formulate this problem as a Machine Learning problem where we predict if a car is in normal mode or in abnormal mode  of operation by considering the OBD CAN bus messages generated by the following ECU’s: Engine Control Module, Electronic Brake control module, Transmission Control Module, Body Control Module, Telematics, Radio, etc.

\section{Methodology}

\begin{figure}[htbp]
			\centering
			\includegraphics[width=0.5\textwidth]{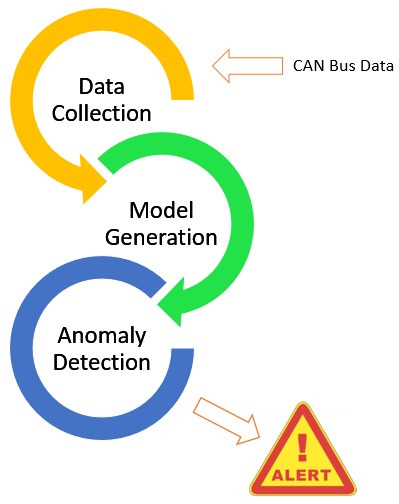}
			\caption{Proposed Detection Architecture}			\label{fig:method}
\end{figure}

We try to convert the problem of detecting abnormal states in a vehicle into a data analytic problem. To accomplish it, we follow the steps as described in Figure~\ref{fig:method}. The first step is the data collection phase in which the stream of CAN bus data is collected. We can employ the OBD-II port present in most vehicles for this purpose. Detailed discussion on data collection can be found in Section~\ref{section:datacollection} The next step is to generate a model which can detect anomalous states in a vehicle. Since Hidden Markov Models (HMM) can abstract the time series data, we use them to model this scenario. Fitting the current scenario to HMM model is described in detail in Section~\ref{section:modelgen}. The final step is the anomaly detection using the generated model. Using HMM's, we can find the posterior probability of a given sequence of observations. Whenever new observations are available, we detect the posterior probability of the current sequence and if the probability falls below a threshold, it indicates a deviation from the normal state of the vehicle and hence we generates an alert in that case. 

\subsection{System Integration}
Our model can be integrated with all current and future car systems as a plug-n-play device or as a system module programmed on the on-board car computer. We can add our system to old cars using their pre-existing OBD Port and attaching a small Rasberry-Pie chip to the OBD port to collect, analyze, and issue alerts. New cars can have this feature pre-installed.

\section{Data Collection}\label{section:datacollection}
\begin{figure}[htbp]
			\centering
			\includegraphics[width=0.5\textwidth]{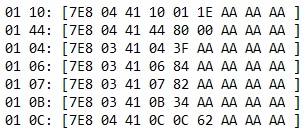}
			\caption{CAN Message}			
			\label{fig:canmessage}
\end{figure}

The first step is to collect the data from the CAN bus. The CAN bus is a broadcast bus on which multiple devices are connected. When a device want to communicate with other components connected to bus, the device will broadcast a message on to the bus with a specific message ID. As shown in Figure~\ref{fig:canmessage}, each CAN message will have a specific Message ID and the message data. While sending the message, each device will be identified by the Message ID alone. It should be noted that all devices connected to the CAN bus will receive the message, but at the receiver end, only those devices which requires that message will accept it while others will just ignore it. To avoid two devices broadcasting together, so that two messages wont be mixed up, It uses a priority mechanism associated with each of the message ID's. 

	It should be noted that the OBD port, which is mandatory in many countries is also connected to the CAN bus in order to collect diagnostic information. Hence we can essentially attach a device on to the OBD-II port and extract data for analysis. There are multiple tools like OBDLink Mx, Blue driver, CAN-BUS Shield which can be connected to Arduino board and ELM 327 clone devices which can be attached to OBD port to extract the raw messages broad-casted over it. 
	
	For data collection, we used STN1100 based OBDLink MX. STN1100 is a multi protocol OBD to UART interpreter integrated circuit. It has a 16 bit processor with inbuilt flash memory and a RAM. It supports the complete AT command set (Command set for ELM 327 based chip-set) along with a new set of ST commands. It supports different protocols like ISO 15765-4 (CAN), ISO 11898 (raw CAN) and SAE J1939 (heavy vehicles). Other selected features include voltage input for battery monitoring and automatic protocol detection. We used OBDWiz, a tool which connect with OBDLink MX, to interface with the vehicles OBD port. During data collection, we set the ``STMA'' command, which will extract all the data flowing over the CAN bus. We collected data from vehicle from different manufacturers which include ``Honda Accord'', ``Toyota Corolla'' and ``Chevrolet Cruze''. 

We faced some practical limitations for collecting the data. Many of these vehicle manufacturers have different mechanisms which hinders direct collection of data from the CAN bus. Some of the techniques include using multiple CAN buses which are guarded by different gateways. These gateways can be unlocked only by specific tools. But these simple techniques wont stop a malicious attacker to crack into it. We were able to collect the information from sensors like  Vehicle speed, load, engine coolant temperature, Engine RPM, Intake air temperature, Absolute throttle position and O2 voltage using the tools mentioned.

\section{Model Generation}
\label{section:modelgen}

\begin{figure}[htbp]
			\centering
			\includegraphics[scale=0.6]{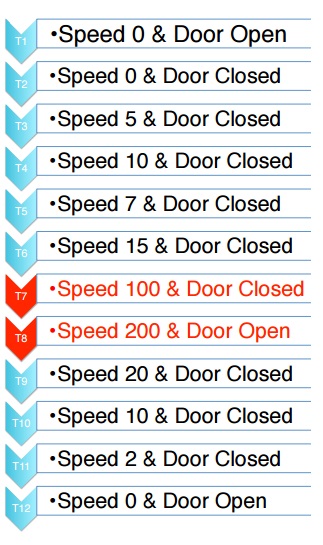}
			\caption{Sample Car Event Time-line}
			\label{fig:intuition}
		\end{figure}

The second step in our approach is to analyze the collected data to develop a model which can identify anomalous states. In this project we try to use Hidden Markov Models (HMM) to create a model. The intuition behind using this model is described below. We consider the movement of a vehicle is nothing but a sequence of states which are dependent on the previous state, like the Markov's processes. For example consider the sequence of activities from $T_{1}$ to $T_{12}$ as shown in Figure~\ref{fig:intuition}. At $T_{1}$ speed is zero and the Door is open. At $T_{2}$ the door is closed and it starts moving. The car gathers speed gradually till $T_{6}$. But at $T_{7}$ there is a sudden jump of 85 miles per hour making the speed to 100 mph. At $T_{8}$ the speed of car is 200 miles per hour and the door is open. We can clearly see that the probability of a state change from $T_{6}$ to $T_{7}$ and $T_{7}$ to $T_{8}$ are unusual. Hence from these sequences we can detect anomalous behaviors.

In order to generate the model, we first separated the data collected into data from different components using custom scripts. To create a model, we used the HMM tool box from Matlab. It has built in functions to perform various HMM operations with high efficiency. The important functions available in it are the following. 
\begin{itemize}
\item hmmgenerate: This function generates a random sequence of observations corresponding to a HMM model which includes the transition and emission probabilities
\item hmmestimate: Given a sequence of observations and states corresponding to it, this function will estimate the corresponding transition and emission probabilities. 
\item hmmtrain: For a given sequence of observations, this function will find the maximum likely hood estimate for emission and transition probabilities. 
\item hmmviterbi: This function determines the most possible sequence of states corresponding to a sequence of observations. 
\item hmmdecode: For a given model and a sequence of observations, this function will estimate the posterior probabilities for the given sequence of observations.

\end{itemize}

The first issue for model generation is how to convert the collected data into a series of observations. For example consider the speed which is collected from the CAN bus. From our collected data we noticed that we have more data at the lower speed bands (0 mph - 10 mph) and average speed bands (40 mph - 60 mph). For other regions the data is really sparse. The problem with such a data model is that it can create a lot of false positives, since we don't have enough data for many state transitions. Hence we trained our model using gradients for the continuous data on the CAN bus. For example in case of speed, instead of using actual speed, we find the speed gradients and train our system for it. The next issue is on how to accommodate multiple observations as a single vector. We have different type of sensors in a vehicular system. Some of them will push data on to CAN bus at regular interval like speed and RPM. On the other hand there are some other observations which are pushed on to the system only when they are required like door sensors in some vehicles. In our model, we create a vector containing different inputs from different systems. Each vector will then represent a single observation and hence our system will be trained for each of those observations.

To train our model, the data interpreted from the CAN bus are represented  as a sequence of observation as mentioned above. For example Speed is 20 mph, RPM is 3000, State of door is closed etc. are modeled as a vector sequence. During implementation, we interpret the different values from particular slots in the CAN message and convert into decimal values from Hex values before using them to train our model. Now these interpreted values are used to generate the sequence of observations. We use this generated sequence of observations to train  HMM model using the ``hmmtrain'' function in Matlab. We chose to use the Baum-Welch algorithm for training which will generate Transition and Emission Probabilities corresponding to test sequences.  

\section{Anomaly Detection}\label{arch:anomaly}
Now the next step is to find any anomalous states. As we described earlier, we are not only detecting attack states, but also any unsafe or anomalous states. For example, even though it is not caused by an attacker, opening of door at 200 mph is unsafe and hence we flag it.
	\begin{figure}[htbp]
			\centering
			\includegraphics[width=0.45\textwidth]{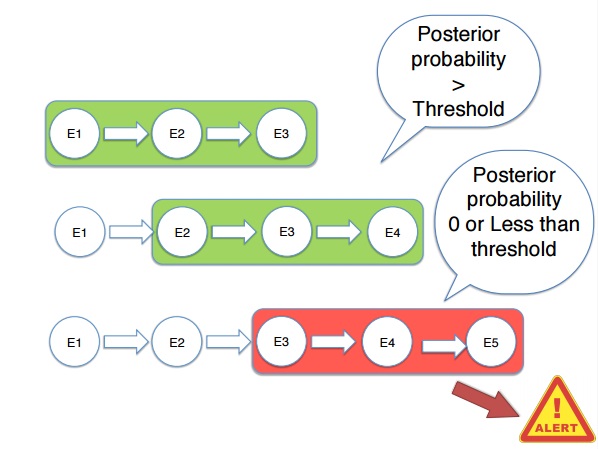}
			\caption{Sliding Window For Anomaly Detection}\label{fig:slidingwindow}
		\end{figure}
To detect unsafe states, we use a sliding window of ``n'' previous observations as shown in Figure~\ref{fig:slidingwindow}. The sliding window moves every time a new observation is available. One of the operations which we can do with HMM is to detect the posterior probability of a given sequence. In this case, once the sliding window is determined, we use all observations in that window and determine the posterior probability of that sequence. In our case each of the observation would be a vector of different sensor values. It will generate a set of probabilities corresponding to each observation. If the probability of any such sequence is below a threshold, based on the generated model, it implies that getting that observation in that sequence is very low and hence we identify it as an anomalous state. 
	
We implemented anomaly detection using the matlab tool box. The anomaly detection module has the model as its first input. In our implementation, the input stream from the CAN bus is fed to this module. It will convert it into a sequences of observation using the same procedure we had used during model generation phase. Now when new observations are available, the module will pick up ``n'' previous observations from the sliding window and use ``hmmdecode'' from matlab to find the posterior probability for the sequence in the window. The module will now generate an alert, if the probability of any observation in the sequence is going below a set threshold value. 

\section{Evaluation \& Results}

For our evaluation, we need to verify that no alerts are generated during normal conditions and alerts are generated during unsafe conditions. To test the normal conditions, we split the collected data into two parts. The first part is used for training the model and the second part is used to verify if the model generates any false positives. For evaluating the system to detect unsafe states, we hand crafted different scenarios by injecting unsafe data into the actual data. We had done a progressive evaluation scheme to test the performance of our model. Our first evaluation used only data from a single sensor. Further evaluations use more than one values at the same time. We describe our evaluation method and corresponding results below. 
\subsection{Single Observation Evaluation}

%------------------------------------------------------
\begin{figure*}[htbp]
 
\begin{subfigure}{0.5\textwidth}
\includegraphics[width=\linewidth]{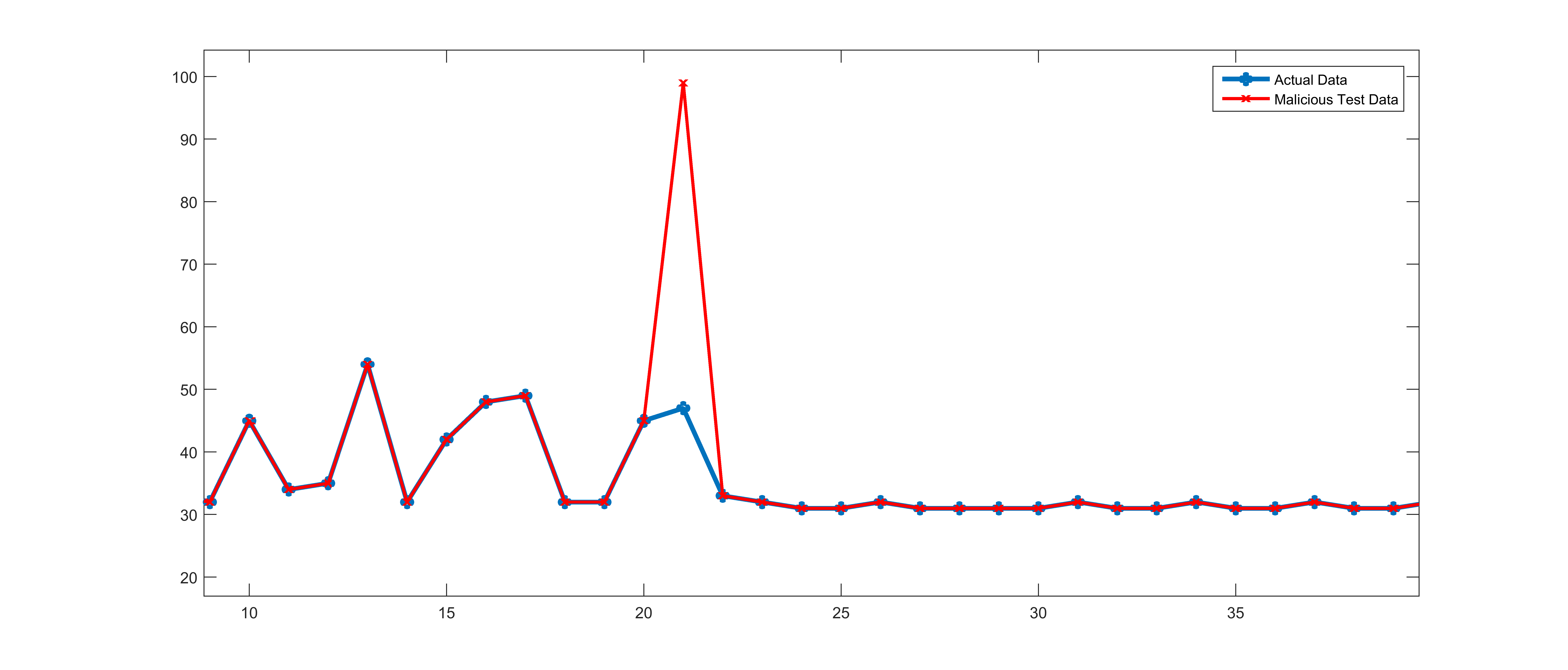} 
\caption{Anomalous RPM Increase}
\label{fig:rsi}
\end{subfigure}
\begin{subfigure}{0.5\textwidth}
\includegraphics[width=\linewidth]{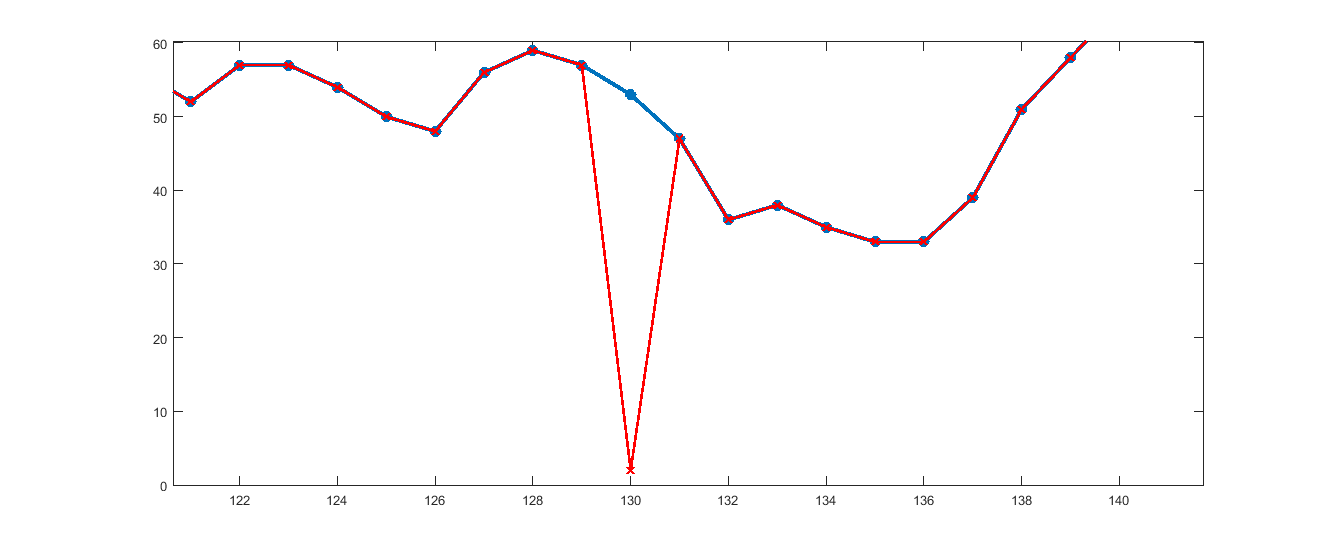}
\caption{Anomalous RPM Decrease}
\label{fig:rsd}
\end{subfigure}
\begin{subfigure}{\textwidth}
\includegraphics[width=0.85\linewidth]{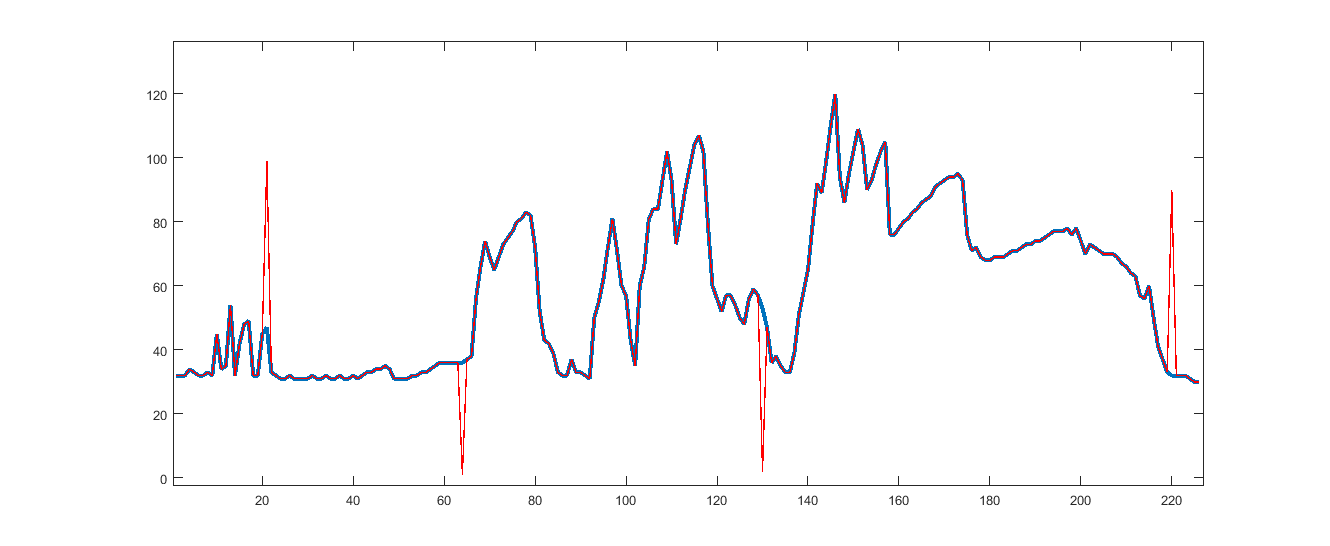}
\label{fig:rfd}
\end{subfigure}
 
\caption{Test Data for RPM as a single observation}
\label{fig:rpmsingle}
\end{figure*}
%------------------------------------------------------
%------------------------------------------------------
\begin{figure*}[htbp]
 
\begin{subfigure}{0.5\textwidth}
\includegraphics[width=\linewidth]{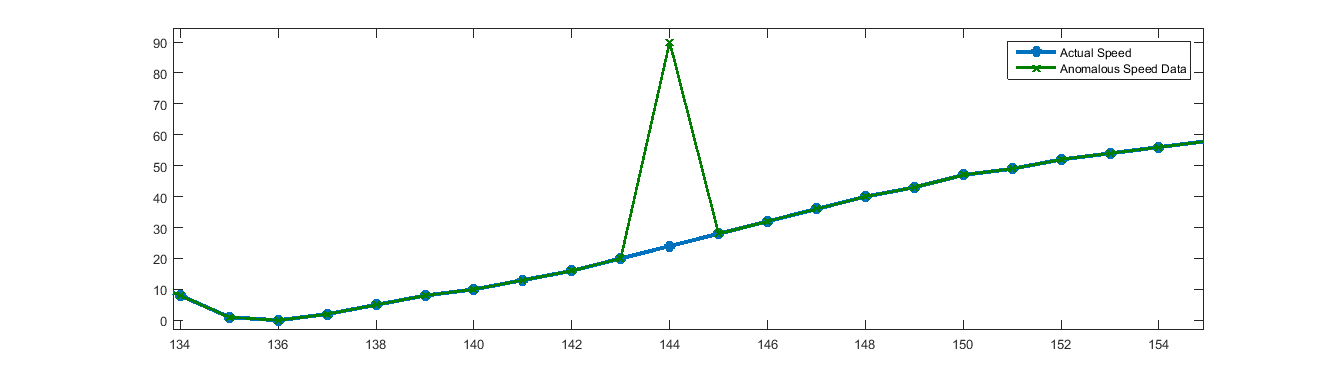} 
\caption{Anomalous Speed Increase}
\label{fig:ssi}
\end{subfigure}
\begin{subfigure}{0.5\textwidth}
\includegraphics[width=\linewidth]{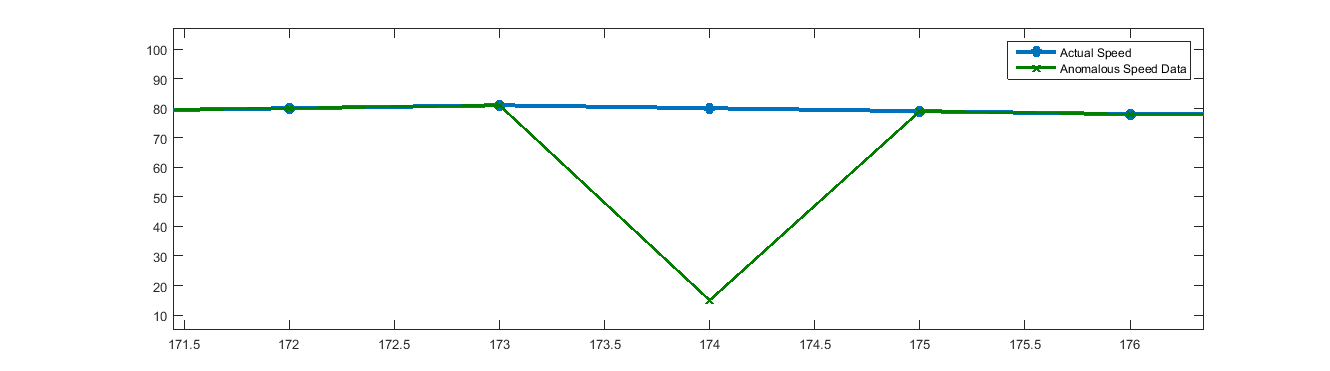}
\caption{Anomalous Speed Decrease}
\label{fig:ssd}
\end{subfigure}
\begin{subfigure}{\textwidth}
\includegraphics[width=0.85\linewidth]{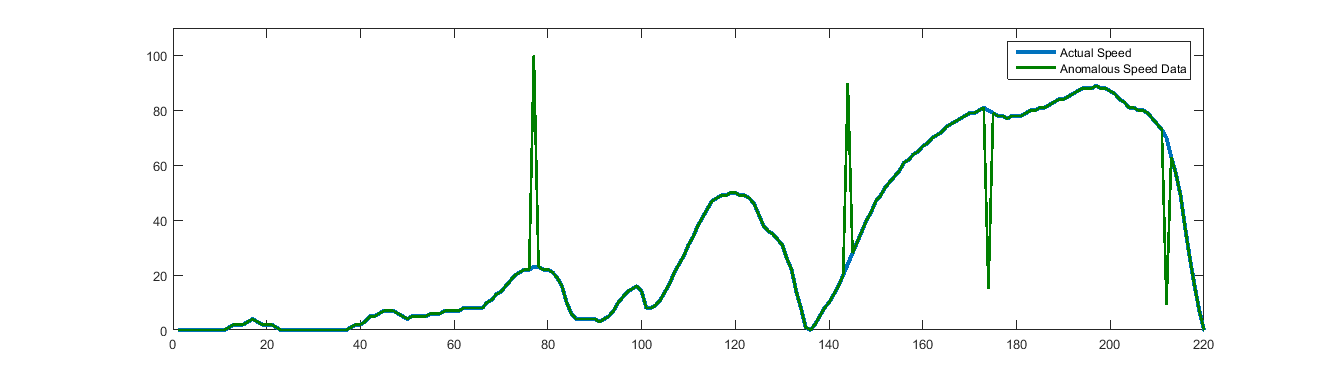}
%\caption{Caption 2}
\label{fig:sfd}
\end{subfigure}
 
\caption{Test Data for Speed as a single observation}
\label{fig:speedsingle}
\end{figure*}
%------------------------------------------------------

%------------------------------------------------------
\begin{figure*}[htbp]
 
\begin{subfigure}{0.24\textwidth}
\includegraphics[width=\linewidth]{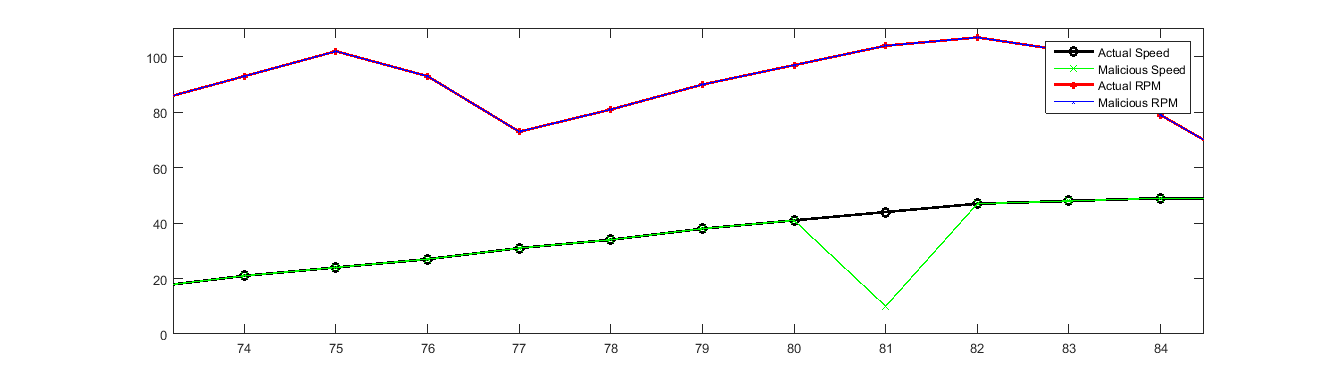} 
\caption{Anomalous Speed Decrease}
\label{fig:msd}
\end{subfigure}
\begin{subfigure}{0.24\textwidth}
\includegraphics[width=\linewidth]{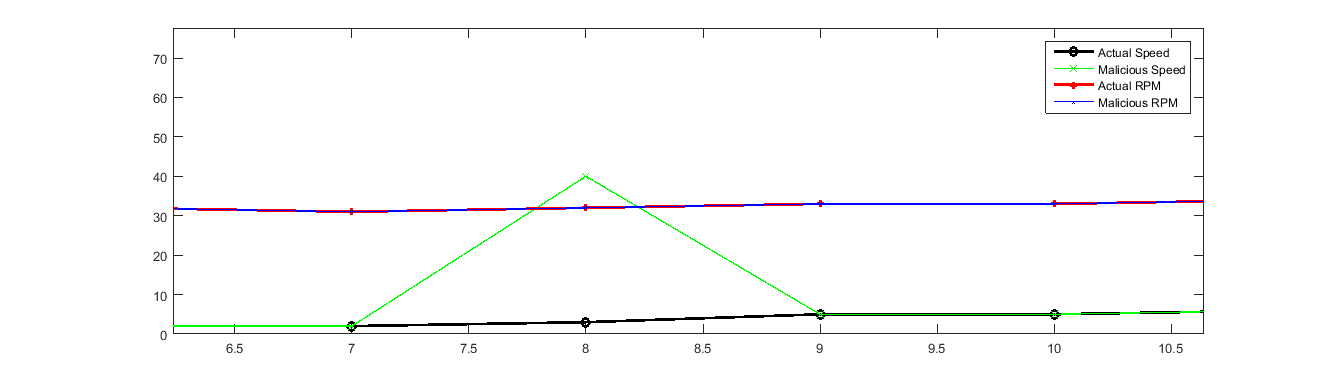}
\caption{Anomalous Speed Increase}
\label{fig:msi}
\end{subfigure}
\begin{subfigure}{0.24\textwidth}
\includegraphics[width=\linewidth]{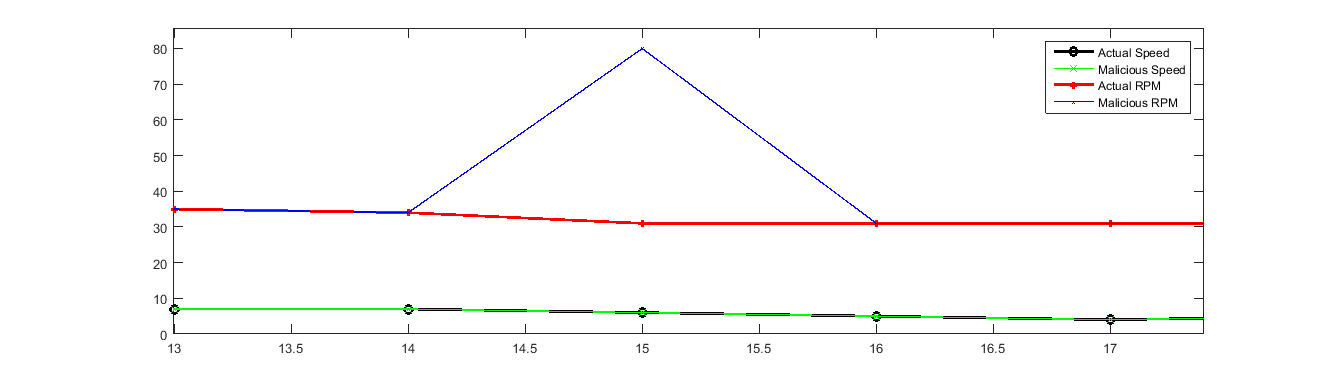}
\caption{Anomalous RPM Increase}
\label{fig:mri}
\end{subfigure}
\begin{subfigure}{0.24\textwidth}
\includegraphics[width=\linewidth]{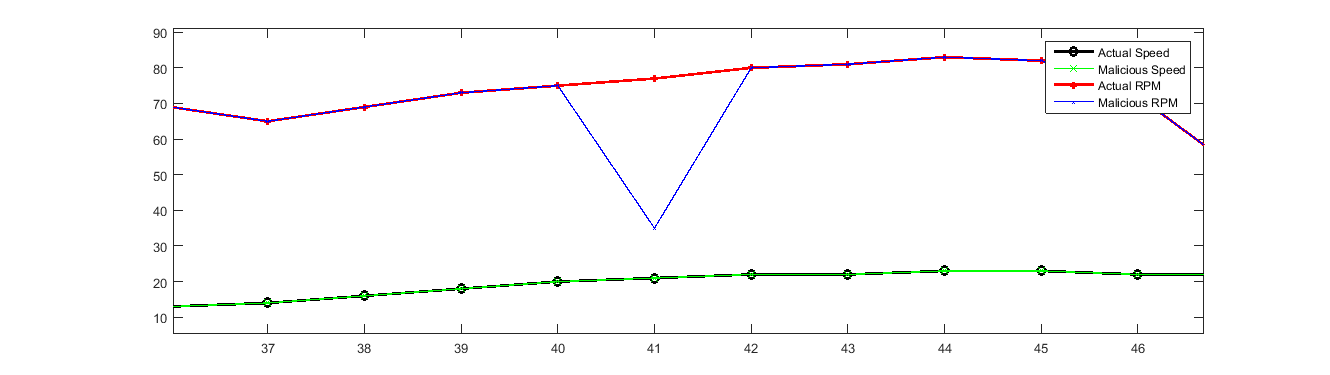} 
\caption{Anomalous RPM Decrease}
\label{fig:mrd}
\end{subfigure}

\begin{subfigure}{0.24\textwidth}
\includegraphics[width=\linewidth]{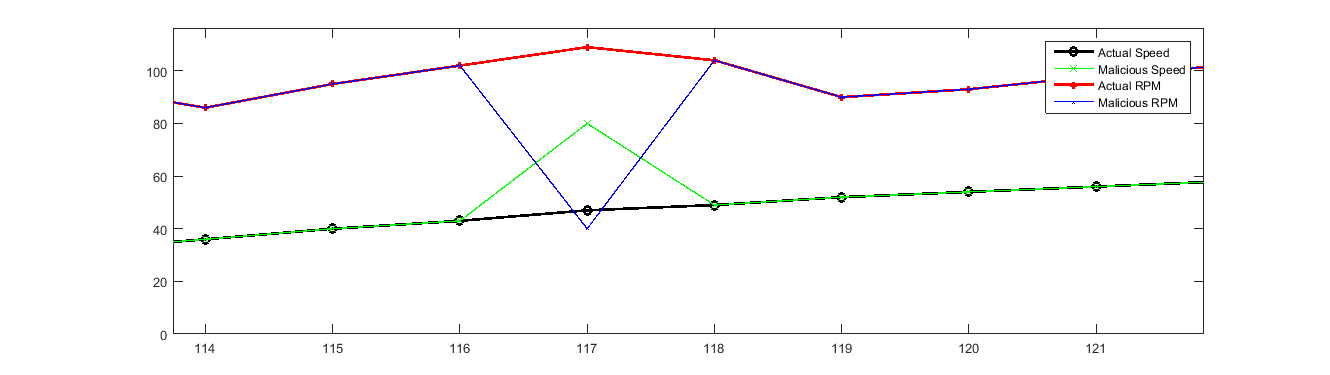}
\caption{Anomalous Speed Increase RPM Decrease}
\label{fig:msird}
\end{subfigure}
\begin{subfigure}{0.24\textwidth}
\includegraphics[width=\linewidth]{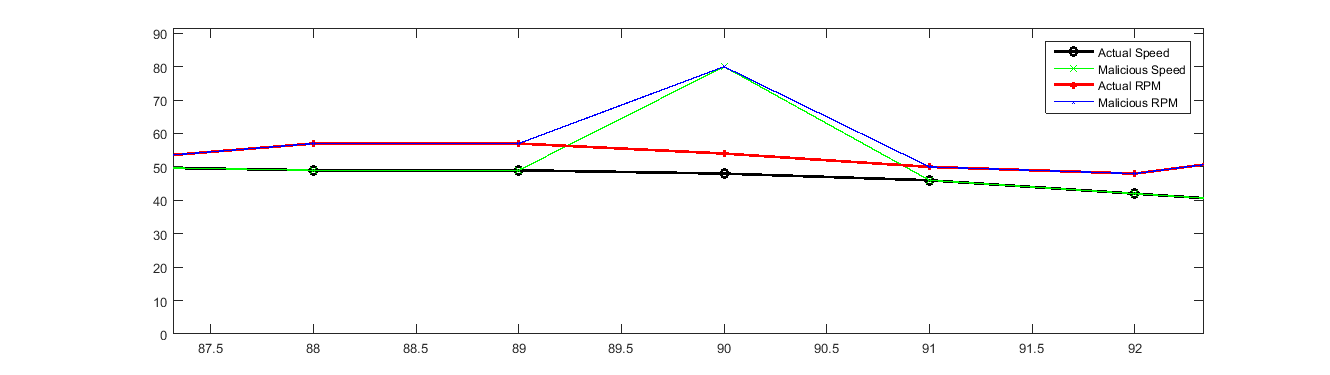}
\caption{Anomalous Speed Increase RPM Increase}
\label{fig:msiri}
\end{subfigure}
\begin{subfigure}{0.24\textwidth}
\includegraphics[width=\linewidth]{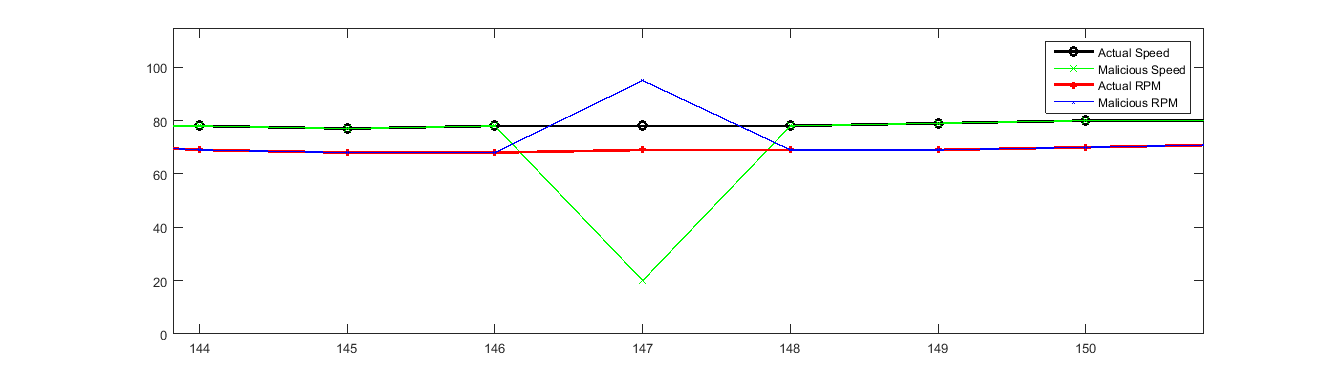}
\caption{Anomalous Speed Decrease RPM Increase}
\label{fig:msdri}
\end{subfigure}
\begin{subfigure}{0.24\textwidth}
\includegraphics[width=\linewidth]{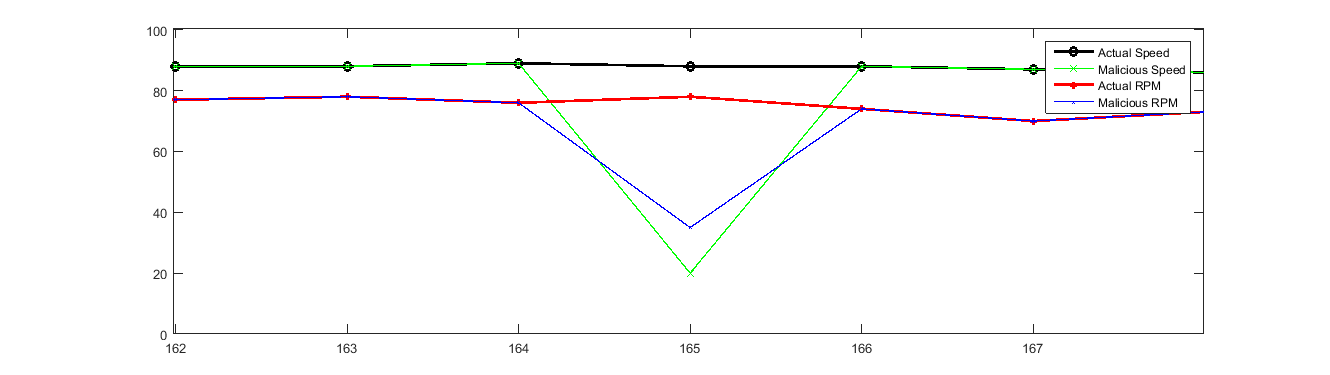}
\caption{Anomalous Speed Decrease RPM Decrease}
\label{fig:msdrd}
\end{subfigure}

\begin{subfigure}{\textwidth}
\centering
\includegraphics[width=0.85\linewidth]{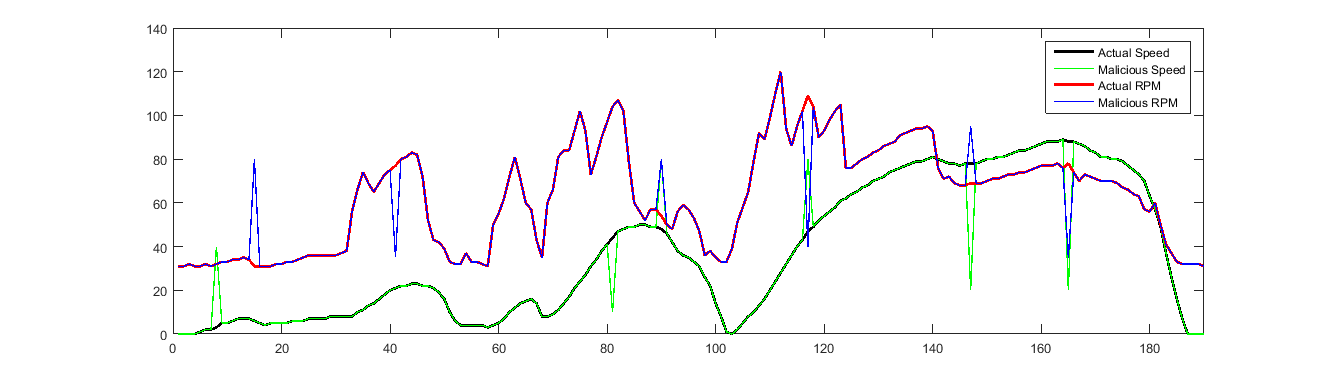}
%\caption{Caption 2}
\label{fig:srfd}
\end{subfigure}
 
\caption{Test Data for RPM and Speed together}
\label{fig:speedrpmmulti}
\end{figure*}
%------------------------------------------------------
We first trained our system only based on a single observed value. We used the data from speed sensor and RPM sensor separately for this. Figure~\ref{fig:speedsingle} represents a part of test data for speed shown graphically. Each of the spikes in it represents the anomalous sudden change in speed caused as a result of the introduction of anomalous data to the real data collected from vehicle. Ideally such a sudden spike is an unsafe state according to our hypothesis. Our generated model was able to detect each of those spikes. In order to make sure that this will work not only for that particular observation, we tried it with RPM sensor data shown in Figure~\ref{fig:rpmsingle}. In a similar way the spike represents a very sudden change in RPM. We should note that the rate of change in RPM and Speed are different. RPM can increase more rapidly than speed. But since our model is based on real data collected from vehicles, it can detect all those variation which will normally happen in them. The results are concluded in the table~\ref{tab:singleobsresults}. We can see that all different anomalous changes, which cannot correspond to the normal context of a car was detected by our generated model. Moreover at those places which do not have spikes the model did not generate any alert.

%\begin{figure*}[htbp]
%			\centering
%			\includegraphics[scale=0.6]{Images/evaluation/eval2.jpg}
%			\caption{Single Observation Evaluation Results}
%			\label{tab:singleobsresults}
%		\end{figure*}

\begin{table}[htbp]
\centering

\begin{tabular}{c|c|c|c|c|c}
\toprule
\rowcolor[HTML]{EFEFEF} 
\cellcolor[HTML]{EFEFEF}                   No& Type&     Speed& \cellcolor[HTML]{EFEFEF}                   Result&  RPM& \cellcolor[HTML]{EFEFEF} Result                  \\
\rowcolor[HTML]{EFEFEF} 
\cellcolor[HTML]{EFEFEF} & of Change&     Alert Status& \cellcolor[HTML]{EFEFEF} &  Alert Status& \cellcolor[HTML]{EFEFEF} \\ \midrule
 1&  $\Uparrow$&              False&      \checkmark&  False \checkmark  \\ \midrule
 2&  $\Downarrow$&            False&      \checkmark&  False \checkmark  \\ \midrule
 3&  $\Uparrow$$\Uparrow$&      True&       \checkmark&  True  \checkmark  \\ \midrule
 4&  $\Downarrow$$\Downarrow$&  True&       \checkmark&  True  \checkmark  \\ \midrule
 5&  $\Longleftrightarrow$&   False&      \checkmark&  False \checkmark  \\ \bottomrule
\end{tabular}
\caption{Single Observation Evaluation. $\Uparrow$ - Gradual Increase, $\Uparrow$$\Uparrow$ - Sudden Increase, \\$\Downarrow$ - Gradual Decrease $\Downarrow$$\Downarrow$ - Sudden Decrease,\\ $\Longleftrightarrow$ - Normal (from Test data)}
\label{tab:singleobsresults}
\end{table}

\subsection{Multiple Observations Evaluation}
Since our model work well with single observations, we evaluated how it will work while considering multiple observations together in a vector. For this evaluation we chose the speed and RPM observations together as a single vector. Both speed and RPM values are generated at regular intervals and hence we could map every speed value with an RPM value. A part of the anomalous values we generated and tested using our model is represented in figure~\ref{fig:speedrpmmulti}. The different spikes represent different anomalous situations which should not happen normally in a vehicle. We tested eight different anomalous situations in it. Each one represents either one of the quantity or both of them being modified which represent a potential malicious state. For example figure~\ref{fig:msird} is the situation in which the speed is suddenly increased while the RPM value suddenly decrease, which is a scenario that can never happen in a normal running vehicle. Similarly figure~\ref{fig:msiri} represents the situation in which the RPM and speed increase rapidly such that it is not physically possible in a normal vehicle. After generating the model, we tested these different cases and the evaluation results are described in table~\ref{tab:multiobsresults}. We can see that the results are promising and could detect such scenarios. But we acknowledge the fact that we need to test our method with more anomalous states of varying degrees. 

\begin{table}[htbp]
\centering

\begin{tabular}{c|c|c|c|c}
\toprule
\rowcolor[HTML]{EFEFEF} 
\cellcolor[HTML]{EFEFEF}                   No& Speed&     RPM& \cellcolor[HTML]{EFEFEF}                   Alert Status&  Result \\ \midrule
1&  $\Uparrow$$\Uparrow$&       $\Uparrow$$\Uparrow$&               True&       \checkmark    \\ \midrule
2&  $\Uparrow$$\Uparrow$&       $\Downarrow$$\Downarrow$&           True&       \checkmark    \\ \midrule
3&  $\Downarrow$$\Downarrow$&   $\Uparrow$$\Uparrow$&               True&       \checkmark    \\ \midrule
4&  $\Downarrow$$\Downarrow$&   $\Downarrow$$\Downarrow$&           True&       \checkmark    \\ \midrule
5&  $\Uparrow$$\Uparrow$&       $\Longleftrightarrow$&            True&       \checkmark    \\ \midrule
6&  $\Downarrow$$\Downarrow$&   $\Longleftrightarrow$&            True&       \checkmark    \\ \midrule
7&  $\Longleftrightarrow$&    $\Uparrow$$\Uparrow$&               True&       \checkmark    \\ \midrule
8&  $\Longleftrightarrow$&    $\Downarrow$$\Downarrow$&           True&       \checkmark    \\ \midrule
9&  $\Longleftrightarrow$&    $\Longleftrightarrow$&            False&      \checkmark    \\ \midrule
\end{tabular}
\caption{Multiple Observation Evaluation. $\Uparrow$$\Uparrow$ - Sudden Increase, $\Downarrow$$\Downarrow$ - Sudden Decrease,\\ $\Longleftrightarrow$ - Normal (from Test data)}
\label{tab:multiobsresults}
\end{table}

\section{Conclusion}

In this project we looked at various security hacks that have surfaced in the recent past. We also studied the CAN Data Bus Model and how to extract data from a CAN Bus. We successfully extracted data from various Ford, Toyota and Honda Cars. Even though data is generated fast and could be collected in bulk from these vehicles, we found that to tackle different issues in this area using data analytic approach, we not only need the data in quantity, we require data with variety also. Hence we identified the requirement of better data sets with variety in this domain for future research activities. Using the collected dataset we generated a Hidden Markov Model for the prediction of anomalous or unsafe states. Our initial results show that such data analytic techniques could be successfully used to identify anomalies and hence unsafe states in a vehicle. Unlike various other methods, such a method could successfully be utilized in both older and newer vehicles. Such techniques could not only protect them from malicious attacks, but also could be employed to assist the driver in various ways. For example such analytics could be used to detect engine failures early or different sensor mal-functioning. 

%\section{Future Work}

%Possible extensions to this project are application of Conditional Random Fields to this problem. We can also move in the direction of a problem specific knowledge graph artificial intelligence based approach to reason over collected data and release alerts based on predetermined policy rules.

\section*{Acknowledgment}

This work is done as a part of the Insure project sponsored by National Science Foundation (NSF). We thank Dr. Alan Sherman for his valuable comments and suggestions during this work. We also thank Mike Moore, Alan Barker and Joesph Raetono from the Oak Ridge National Laboratory, for answering our various queries related to practical issues, we faced during this work.

\bibliographystyle{plain} 
\bibliography{ref.bib}

\end{document}